\documentclass[runningheads]{llncs}

\usepackage{eccv}

\usepackage{eccvabbrv}
\usepackage{amssymb}
\usepackage{pifont}
\newcommand{\cmark}{\checkmark}
\newcommand{\xmark}{\ding{55}}

\usepackage{graphicx}
\usepackage{booktabs}

\usepackage{multirow}
\usepackage{subcaption}
\usepackage[table]{xcolor}
\usepackage{placeins}
\usepackage[ruled,vlined]{algorithm2e}
\usepackage{tabularx} 

\usepackage{makecell}

\usepackage{xcolor}
\newcommand{\std}[1]{\,{\scriptsize\color{gray}({#1})}}
\usepackage[accsupp]{axessibility}  

\usepackage{hyperref}

\usepackage{orcidlink}

\begin{document}

\title{Navigating Gigapixel Pathology Images with Large Multimodal Models} 
\titlerunning{Navigating Gigapixel Pathology Images with Large Multimodal Models}

\author{Thomas A. Buckley\inst{1*} \and
Kian R. Weihrauch\inst{1*} \and
Katherine Latham\inst{2}\and \\ Andrew Z. Zhou\inst{1}\and Padmini A. Manrai\inst{3}\and Arjun K. Manrai\inst{1}}

\institute{Department of Biomedical Informatics, Harvard Medical School \and
Department of Pathology, Massachusetts General Hospital \and Department of Pathology and Laboratory Medicine, Brown University}

\authorrunning{Buckley, Weihrauch et al.}

\maketitle
\def\thefootnote{*}\footnotetext{These authors contributed equally.}
\def\thefootnote{}\footnotetext{Contact: Arjun\_Manrai@hms.harvard.edu}

\begin{abstract}
Recent advances in large multimodal models have allowed for the development of interactive 
chat models that can converse and reason about pathology whole-slide images (WSIs).
However, existing slide-level chat systems are often highly specialized, typically compressing WSIs into fixed slide-level embeddings or relying on multi-component pipelines, which can lose multi-scale detail and limit generalizability beyond the target task.
We present \textbf{GIANT} (\textbf{G}igapixel \textbf{I}mage \textbf{A}gent for \textbf{N}avigating \textbf{T}issue), a simple, training-free approach that lets general-purpose multimodal models navigate WSIs on their own, iteratively selecting multi-magnification crops and aggregating evidence over time.
To evaluate generalizability in WSI question answering and to promote reproducibility, we introduce \textbf{MultiPathQA}, a benchmark suite spanning five clinical challenges and 934 questions over 868 unique WSIs. This includes a new set of 128 pathologist-authored multiple-choice questions designed to mirror real diagnostic search and multi-scale reasoning.
Using GPT-5, GIANT outperforms models specialized for pathology question answering, achieving state-of-the-art performance on four out of five benchmarks. 
\end{abstract}

\section{Introduction}
\label{sec:intro}

\begin{figure}[t]
    \centering
    \includegraphics[width=0.8\linewidth]{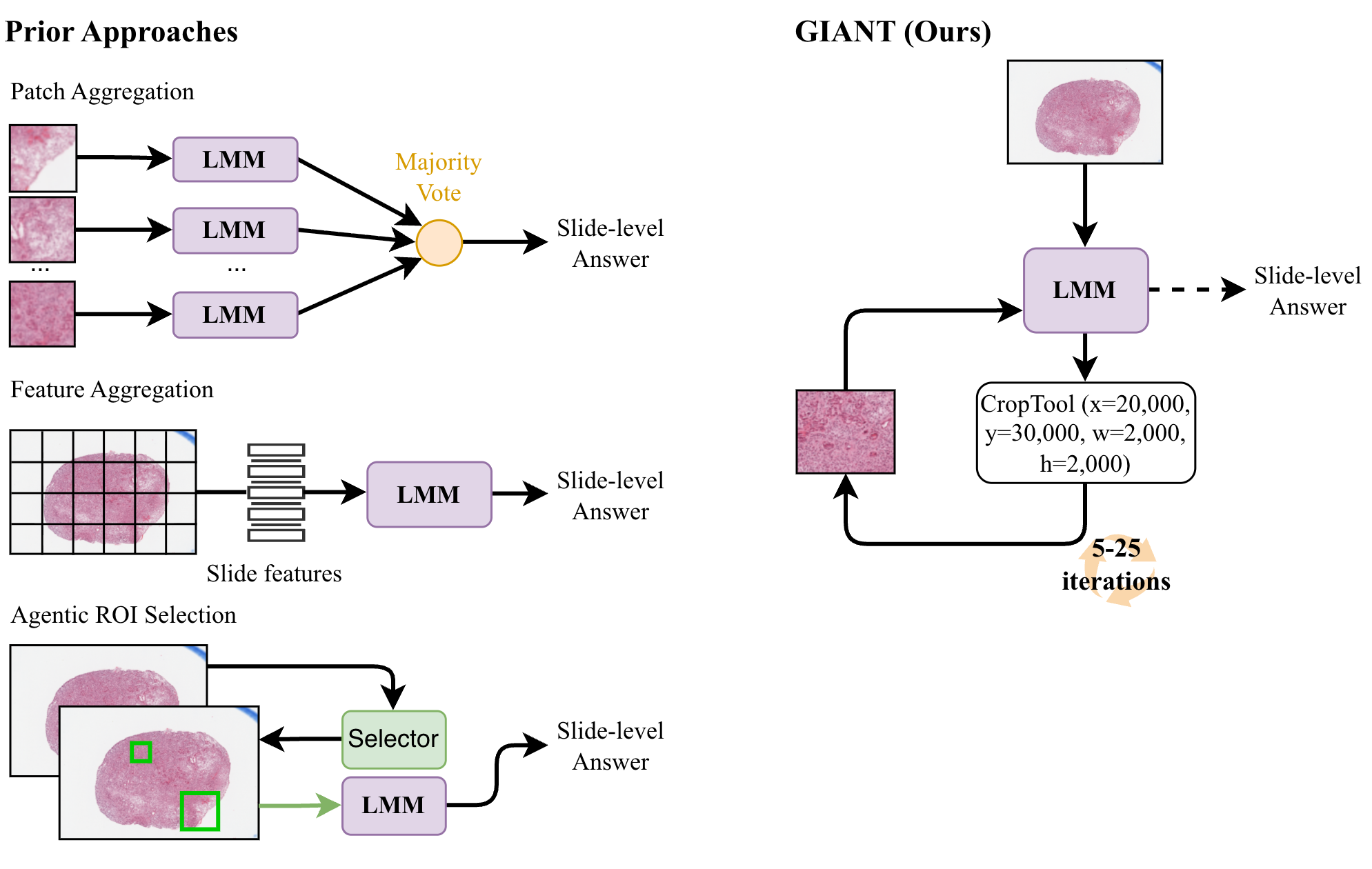}

    \caption{In this study, we introduce \textbf{G}igapixel \textbf{I}mage \textbf{A}gent for \textbf{N}avigating \textbf{T}issue (\textbf{GIANT}), a framework that enables general-purpose LMMs to iteratively pan, zoom, and reason across a gigapixel pathology image. We contrast GIANT with prior strategies that use random patch sets, learned slide-level embeddings, or selector-based agentic pipelines for slide-level analysis. We show that our approach outperforms specialized pathology chat models trained on thousands of whole-slide images.}
    \label{fig:overall}
\end{figure}

\begin{figure}[t]
  \centering
  \includegraphics[width=\linewidth]{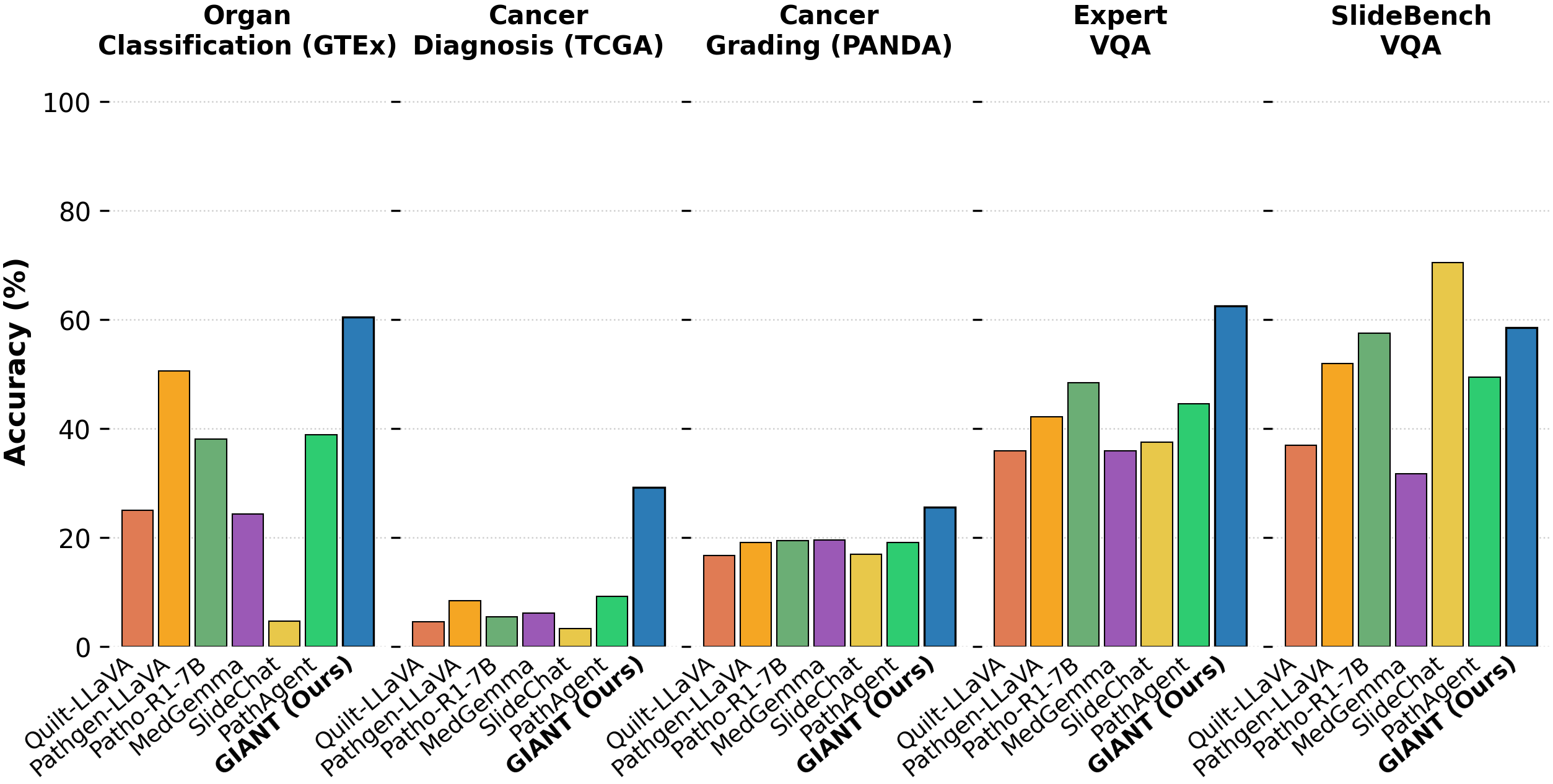}
  \caption{Barplot comparison of the GIANT (GPT-5 x5) approach versus leading chat models trained on whole-slide pathology images. We compare GIANT against slide-level question answering models (SlideChat, MedGemma, PathAgent), and ROI models (Quilt-LLaVA, Pathgen-LLaVA, Patho-R1-7B). For ROI-based models, we choose the best performing input aggregation method from either random patch ensembling or thumbnail.}
  \label{fig:results1}
\end{figure}

\begin{figure}[t]
  \centering
  \includegraphics[width=\linewidth]{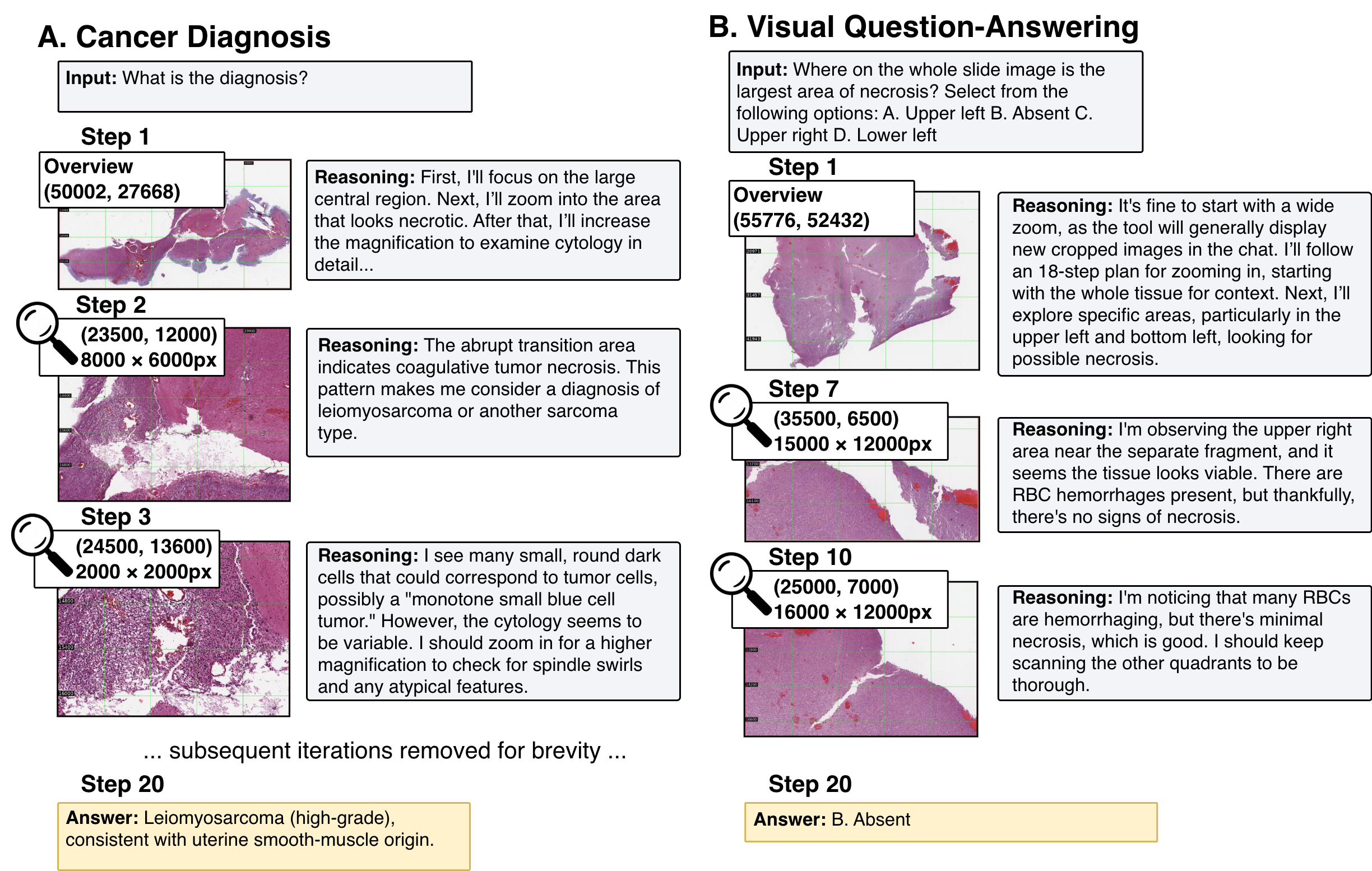}
  \caption{Examples of GIANT reasoning traces for cancer diagnosis and multiple-choice question-answering. The model is instructed to produce its final response after 20 iterations, which can be modified using a system prompt. See Supplementary Material for prompts.}
  \label{fig:short}
\end{figure}

\begin{figure*}[t]
  \centering
  \includegraphics[width=\linewidth]{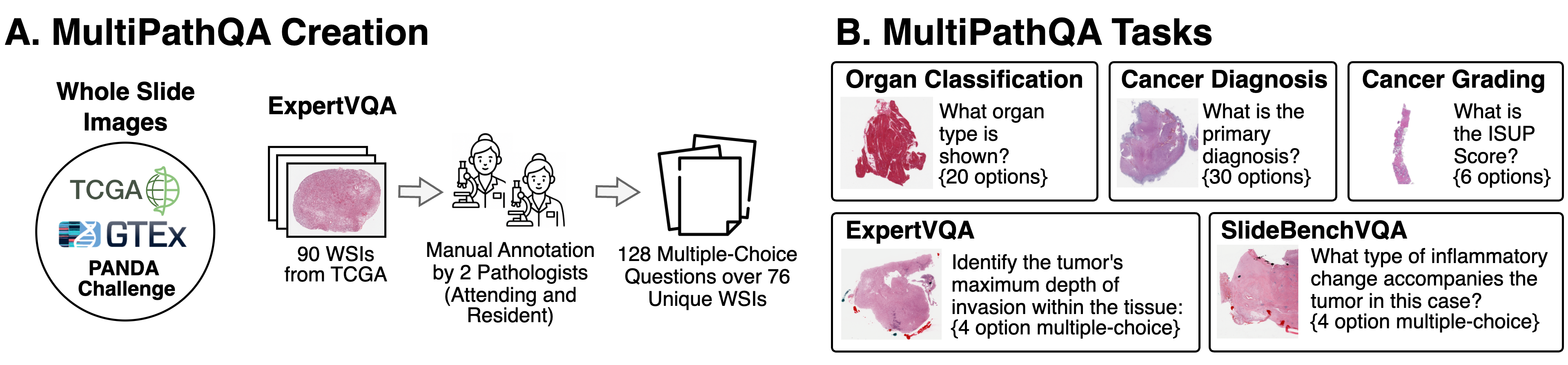}
  \caption{\textbf{A.} We introduce \textbf{MultiPathQA}, a new publicly-available benchmark suite for evaluating LMMs on whole-slide question answering, assessing both diagnostic performance and clinically relevant question-answering capabilities. It uses WSIs from TCGA, GTEX, and the PANDA Challenge. In our study, two pathologists (one resident, one attending) generated 128 questions after manual review of 96 TCGA slides. These questions are designed to require direct image interpretation at multiple scales. \textbf{B.} Examples of the five tasks in MultiPathQA.}
  \label{fig:multipathqa}
\end{figure*}

Asking questions over gigapixel whole slide images is a critical task in medical decision-making, cancer staging, and translational research \cite{Amin2017AJCC8th}.  Recent advances in large multimodal models (LMMs) and vision language models, trained on millions of images, enable free-text question answering on histology images \cite{he2020pathvqa, sun2024pathmmu, chen2024wsivqa, chen2025slidechat, liang2025wsillava, ikezogwo2023quilt, sun2025pathgenm}. These models have shown promise in cancer detection and grading, molecular classification, and survival modeling.
However, almost all existing approaches rely on task-specific training, optimizing the model for a particular benchmark or dataset. This specialization can limit generalizability in real-world pathology settings. Additionally, limited public availability of model weights and datasets hinders reproducibility and comparative evaluation. As a result, the performance of pathology question-answering models across diverse tasks in clinical pathology, and what strategies generalize across tasks remain poorly understood.

In this work, we introduce \textbf{G}igapixel \textbf{I}mage \textbf{A}gent for \textbf{N}avigating \textbf{T}issue (\textbf{GIANT}), a simple and lightweight approach that allows non-specialized LMMs to iteratively pan, zoom, and reason over pathology WSIs. GIANT acts on its own by placing the LMM in the loop, allowing it to actively navigate the image, rather than relying on external models for navigation. 

To systematically evaluate model capabilities in pathology question answering, we introduce \textbf{MultiPathQA}, a benchmark spanning five clinical challenges and 934 questions across 868 unique WSIs. MultiPathQA evaluates a spectrum of tasks, from basic organ identification and organ-cancer classification to cancer staging. The benchmark combines several evaluation tasks with a newly-developed dataset called \textbf{ExpertVQA}, a collection of 128 questions manually created by two practicing pathologists. These expert-authored questions mirror the workflow of human pathologists, testing the model’s ability to integrate contextual and multi-scale information in real-world settings.

Using MultiPathQA, we find that existing pathology-trained WSI chat systems often fail to generalize outside of their training data. This leads to broad underperformance on diverse pathology tasks. We find that GIANT, using GPT-5 as the backbone, substantially outperforms specialist chat models. 

We conduct ablation studies including (1) text-only experiments to measure if our system improves visual reasoning beyond the text, and (2) pathologist experiments to measure the quality of GIANT-chosen navigation strategies.

Overall, our key contributions are:
\begin{enumerate}
\item \textbf{GIANT:} a simple agentic framework that enables training-free visual question answering over gigapixel pathology images. In contrast to prior work, GIANT does not require separate, trained navigation models but navigates the slide to find regions relevant to the prompt with the LMM in the loop. 

\item \textbf{MultiPathQA:} a comprehensive benchmark of 934 pathology questions across 868 WSIs, including ExpertVQA, a new pathologist-created benchmark designed to reflect diagnostic reasoning.

\item \textbf{A systematic evaluation} of generalist and specialist models for pathology question answering across diverse tasks, finding that existing models for pathology question-answering struggle to generalize to new and realistic tasks in pathology.

\end{enumerate}


\section{Related Work}


\begin{table}[!t]
\centering
\small
\setlength{\tabcolsep}{6pt}
\caption{
Comparison of conversational models in pathology.
ROI: region-of-interest level. WSI: whole-slide image level.
}
\begin{tabular}{lcccc}
\toprule
\textbf{Model} & \textbf{Chat} & \textbf{No Fine-tuning} & \textbf{Publicly Available} & \textbf{Level} \\
\midrule
\textbf{GIANT (Ours)} & \cmark & \cmark & \cmark & WSI \\

\midrule
Quilt-LLaVA      & \cmark & \xmark & \cmark & ROI \\
Pathgen-LLaVA    & \cmark & \xmark & \cmark & ROI \\
Patho-R1-7B      & \cmark & \xmark & \cmark & ROI \\
WSI-Agents       & \cmark & \cmark & \xmark & WSI \\
WSI-LLaVA        & \cmark & \xmark & \cmark & WSI \\
PathFinder       & \cmark & \xmark & \xmark & WSI \\
SlideSeek        & \cmark & \xmark & \xmark & WSI \\
PathAgent        & \cmark & \xmark & \cmark & WSI \\
CPathAgent        & \cmark & \xmark & \xmark & WSI \\
SlideChat        & \cmark & \xmark & \cmark & WSI \\
PathChat         & \cmark & \xmark & \xmark & ROI \\
Pathology-CoT    & \cmark & \xmark & \cmark & WSI \\
\bottomrule
\end{tabular}
\label{tab:model_comparison}
\end{table}
GIANT builds off of extensive prior work that has explored multimodal chat systems with pathology images. We categorize current approaches by whether they operate on pre-selected regions of interest (ROIs), or WSIs, and by whether the system depends on fine-tuning using pathology images. 
Table~\ref{tab:model_comparison} provides an overview of existing approaches, and whether each model is publicly available.

\subsection{Fine-Tuned ROI Chat Models}
Pathology-trained LMMs for ROI analysis focus on domain-specific alignment, enhancing the ability of LLMs to interpret morphological features and disease patterns within localized tissue patches. These models typically operate on pre-selected ROIs and therefore do not address the computational challenges of ingesting gigapixel WSIs.
Most approaches follow a similar architecture: a vision encoder, pre-trained on medical image–text pairs, extracts visual embeddings, which are then aligned with a language model through visual instruction tuning. Representative examples include PathChat \cite{Lu2024MultimodalAIPathology}, PathGen-LLaVA \cite{sun2025pathgenm}, Quilt-LLaVA \cite{seyfioglu2025quiltllavavisualinstructiontuning}, and Patho-R1 \cite{zhang2025pathor1multimodalreinforcementlearningbased}. Collectively, these models enable interactive, single-region dialogue, acting as an assistant to a pathologist who may want to discuss a specific region of the slide.

\subsection{Fine-Tuned WSI Chat Models}
Pathology-trained WSI chat models aim to overcome the challenge of ingesting full WSIs to enable holistic slide evaluation. One strategy is to compress the slide into a compact representation. Models such as SlideChat \cite{chen2025slidechat} and WSI-LLaVA \cite{liang2025wsillava} aggregate tissue patches into a unified slide-level embedding. Following a similar motivation, MedGemma 1.5 4B \cite{medgemma} employed a slightly different approach, preserving broader context by passing the embeddings of multiple random low-magnification ROIs to a fine-tuned LMM. However, because these approaches introduce new embeddings, the underlying language model typically must be fine-tuned to accept them, often pushing designs toward smaller backbones for efficiency. 

Another direction constructs WSI-level chat models via agentic, multi-step pipelines that decouple ROI selection from ROI interpretation. In these systems, one component proposes candidate regions (often using low-magnification context), while a second component performs pathology-specific analysis and language generation on the selected ROIs. Representative examples include PathAgent \cite{chen2025pathagentinterpretableanalysiswholeslide}, SlideSeek \cite{chen2025slideseek}, PathFinder \cite{ghezloo2025pathfindermultimodalmultiagentmedical}, Pathology-CoT \cite{wang2025path}, and CPathAgent \cite{sun2025cpathagentagentbasedfoundationmodel}. PathAgent and SlideSeek use a general-purpose model for ROI selection and a pathology-trained LMM for reasoning. PathFinder uses a pathology-trained segmentation model for region selection. Pathology-CoT also trains a task-specific selection model while leveraging general-purpose LMMs for downstream analysis; however, its current selector is limited to lymph-node slides only.

\subsection{Non-specialized Multimodal Chat Models}
General-purpose multimodal language models demonstrate strong cross-domain reasoning and contain substantial medical knowledge\cite{superhuman, brin2023comparing}.
However, they cannot directly ingest gigapixel-sized WSIs. As a result, prior work typically resorts to coarse slide representations, such as random patch sampling or low-resolution thumbnails \cite{chen2025slidechat, chen2025pathagentinterpretableanalysiswholeslide, liang2025wsillava}.
Pathology-CoT leverages a general-purpose LMM for diagnosis within an agentic framework, but it does not allow the LMM to explore the slide. Instead, the model is provided to regions of interest selected by a segmentation model trained for a lymph-node–specific, annotated task.
Code for PathChat, SlideSeek, PathFinder, and CPathAgent is not publicly available.

\subsection{Foundation Models for WSI}
A complementary line of work trains pathology foundation models on large-scale histology datasets. CLIP-style vision--language encoders such as PLIP~\cite{huang2023visual}, Quilt/Quilt-1M~\cite{ikezogwo2023quilt}, CONCH~\cite{lu2024avisionlanguage}, and Virchow~\cite{vorontsov2024virchow} learn patch-level representations aligned with text, enabling strong retrieval and zero-shot classification. Building on such patch encoders, WSI-level models aggregate patch embeddings into slide representations, e.g., PRISM/PRISM2~\cite{shaikovski2024prism,shaikovski2025prism2unlockingmultimodalgeneral} and TITAN~\cite{titan_model}. Vision-only WSI foundation models such as CHIEF~\cite{wang2024pathology} and UNI / Prov-GigaPath~\cite{chen2024towards,xu2024whole} likewise learn powerful slide-level features. These models typically operate on static, precomputed representations and do not support interactive, prompt-driven multi-scale exploration of WSIs, motivating agentic approaches for open-ended question answering.

\subsection{Benchmarks for WSI}

High-quality, publicly available benchmarks for whole-slide VQA remain limited. One benchmark testing differential diagnosis from WSIs, DDxBench, was introduced with SlideSeek~\cite{chen2025slideseek}, but is not publicly available.
Among public benchmarks, SlideBench \cite{chen2025slidechat}, WSI-VQA \cite{chen2024wsivqa}, and WSI-Bench \cite{liang2025wsillava} all generate questions automatically from pathology reports using large language models. SlideBench includes retrospective pathologist validation, whereas the others do not.

Across these datasets, and as noted in prior work, the question-generation pipelines rely on simple templates that often yield low-quality or weakly grounded questions \cite{liang2025wsillava, chen2024wsivqa}. SlideBench is generated with pathologist oversight, and we therefore include it in our evaluation. However, even SlideBench exhibits notable issues. Several WSIs appear to have unusually low resolution or small file sizes, suggesting low-magnification scanning or aggressive downsampling; in one instance, the slide contains a single region of interest rather than a full WSI. Moreover, SlideBench’s question set was derived using the same pipeline used to generate training instructions for SlideChat, adding the possibility that the reported benchmark results for SlideChat may overestimate real-world generalizability. 
These limitations motivated us to develop ExpertVQA, a pathologist-generated VQA benchmark that ensures high image quality and clinically meaningful questions.

\section{MultiPathQA}
\label{sec:benchmark}

We first developed a new benchmark called \textbf{MultiPathQA} for WSI question-answering, comprising five key tasks in pathology as shown in Figure~\ref{fig:multipathqa}. All WSIs were quality-checked by board-certified pathologists in this study or prior work~\cite{gtex_quality_assessment, tcga_ut_benchmark, chen2025slidechat, pandachallenge}. We release the dataset publicly on HuggingFace. See Supplement for dataset statistics.

\subsection{ExpertVQA}

This new pathologist-created dataset consists of multiple-choice questions requiring spatial reasoning over WSIs. WSIs were drawn from the TCGA Uniform cohort~\cite{tcga_ut_benchmark}, which contains 8{,}736 WSIs for 7{,}175 patients across 32 cancer types. All images in this dataset have been screened by pathologists for quality and correctness in a prior study~\cite{tcga_ut_benchmark}. Using stratified sampling across the 20 most frequent primary sites, we selected 90 WSIs (one per patient). Two pathologist co-authors (one attending, one resident) reviewed the available slides and selected a subset for question generation. For the selected slides, they formulated 1-3 diagnostic questions requiring direct image interpretation and search (e.g., “Which region contains necrosis?”). Each item includes four answer choices and one verified correct label.  
In total, the dataset contains 128 questions from 76 WSIs covering 20 diagnostic categories.
Patients included in ExpertVQA were excluded from all other datasets.

\subsection{Organ Classification (GTEx)}

WSIs were obtained from the GTEx project~\cite{gtex}, which provides histology slides from non-diseased donor tissues. All slides underwent quality control for tissue integrity, autolysis, and processing artifacts~\cite{gtex_quality_assessment}. Slides showing severe degradation or extensive pathology were excluded; slides with minor physiological variation (e.g., mild fibrosis or atrophy) were retained. From 25{,}713 entries (980 donors), the 20 most frequent organ types were identified, and stratified sampling by organ type was used to select 200 WSIs. Four slides failed retrieval,  and another five failed segmentation using the CLAM pipeline, yielding 191 final WSIs from 173 donors. Each WSI is annotated with organ site and metadata describing autolysis and histological findings. Evaluation is performed as a 20-way classification task using top-1 balanced accuracy.

\subsection{Cancer Diagnosis (TCGA)}

The cancer diagnosis dataset was derived from the TCGA Uniform cohort~\cite{tcga_ut_benchmark, chen2024towards}. After removing patients appearing in ExpertVQA, WSIs were sampled stratified by cancer type. Following prior work, colon (COAD) and rectal (READ) adenocarcinomas were merged into a single “Colon–Rectum adenocarcinoma” category ~\cite{chen2024towards}. Testicular Germ Cell Tumors (TGCT) images were unavailable and needed to be excluded. This resulted in 221 slides from 197 patients across 30 unique cancer types. Evaluation is performed as a 30-way classification task, reporting top-1 balanced accuracy.

\subsection{Cancer Grading (PANDA)}

The prostate cancer grading dataset was derived from the PANDA challenge collection containing 10{,}616 WSIs with ISUP grade annotations. From this set, 200 slides were selected using stratified sampling across six categories corresponding to ISUP grades 1–5 and benign cases (labeled as grade 0) \cite{pandachallenge}. Three slides failed segmentation using the CLAM pipeline and were excluded, resulting in 197 slides. Evaluation uses grade-level balanced classification accuracy.

\subsection{SlideBenchVQA}

We downloaded SlideChat-VQA-TCGA-plus, published as part of SlideChat, which has 3,176 samples from TCGA across 31 cancer types\cite{chen2025slidechat}. We selected a random sample of 200 questions from this dataset, which we refer to as SlideBenchVQA. Three slides failed segmentation with the CLAM pipeline or were missing, resulting in 197 questions over 183 unique images.

\section{GIANT}
\label{sec:method}





\subsection{Overview}
GIANT turns an LMM into an \textbf{interactive agent} that navigates a WSI through a sequence of image-crop actions. Starting from a low-resolution thumbnail, the agent iteratively selects regions to examine, guided by its evolving interpretation of the slide.
At iteration $t$, the agent observes the crop $I_t$, produces reasoning $r_t$, and proposes the next action $a_t$. The environment returns the next image $I_{t+1} = \text{CropRegion}(W, a_t, S)$, repeating until a maximum iteration limit $T$ (see \ref{sec:design_choices}). The sequence $\{(I_t, r_t, a_t)\}_{t=1}^{T}$ constitutes the multimodal trajectory for that slide. The thumbnail is overlaid with four evenly spaced axis guides along each dimension to orient the model.

\begin{algorithm}[h]
\caption{GIANT}
\label{alg:giant_fusion}
\SetKwInOut{Input}{Input}\SetKwInOut{Output}{Output}
\SetKwProg{Fn}{Function}{}{}

\Fn{\textsc{GIANT}($W, q, T, S$)}{
    \Input{WSI $W \in \mathbb{R}^{H_0 \times W_0 \times C_{img}}$, question $q$, max steps $T$, long-side $S$}
    \Output{Answer $\hat{y}$}
    \BlankLine
    $I_0 \leftarrow \text{Thumbnail}(W)$ \\
    $P_0 =$ \{$q$, nav instructions + \text{``at most }T{-}1\text{ crops''}\}\
    $C \leftarrow [I_0, P_0]$ \\
    \For{$t \leftarrow 1$ \KwTo $T{-}1$}{
        $(r_t, a_t) \leftarrow \text{LMM}(C)$ \tcp*{$a_t=(x,y,w,h)$}
        $I_t \leftarrow \text{CropRegion}(W, a_t, S)$ \\
        $C \leftarrow C \,\Vert\, (r_t, a_t, I_t)$
    }
    $\hat{y} \leftarrow \text{LMM}(C)$ \\
    \Return $\hat{y}$
}

\BlankLine

\end{algorithm}

\subsection{Prompt Development}
All prompts for GIANT are included in the Supplementary Materials. To prevent test-set leakage, we developed all prompts on a held-out set of 10 images before evaluation against our benchmark. Prompts are simple with no task-specific instructions to ensure generalizability (see Supplement).

\subsection{GIANT System Design Choices}
\label{sec:design_choices}
To choose the thumbnail and crop size, we conducted a sweep of performance across thumbnail size, finding that performance quickly saturates after 1000 pixels (see Supplementary Materials). We therefore choose this as our thumbnail and crop size. To choose iteration count, we conducted an experiment sweeping the number of iterations from 1 to 25 in increments of 5, finding that accuracy across all tasks is highest at 20 iterations (see Supplementary Materials). 

\subsection{Baseline Models}
\label{sec:baselines}

We compare GIANT to 10 pathology chat systems, spanning specialized and non-specialized pathology models. See Supplementary Materials for standardized task prompts. 

\paragraph{ROI Chat Baselines:}
For generalist models, we evaluate GPT-5, GPT-4o, and Qwen3-VL-235B. For specialist models, we evaluate Patho-R1-7B, Pathgen-LLaVA, and Quilt-LLaVA. For fair comparison to WSI models, we adopt two standard configurations widely adopted in prior studies~\cite{liang2025wsillava, chen2025slidechat, li2025, chen2025pathagentinterpretableanalysiswholeslide}:
\begin{itemize}
    \item \textbf{Thumbnail baseline:} Each model receives a 1024$\times$1024 thumbnail of the entire WSI using the same resolution as in prior work ~\cite{chen2025slidechat}.  
    \item \textbf{Random patch aggregation:} Following SlideChat~\cite{chen2025slidechat}, we sample 30 random 224$\times$224px patches at 20$\times$ magnification from each WSI after tissue segmentation.  
    The model independently answers each patch, and predictions are combined by majority vote.  
\end{itemize}

For Quilt-LLaVA, the model did not adhere to formatting instructions, so we used an instance of GPT-5-mini to extract the chosen option from the text output (see Supplementary Materials).

\paragraph{WSI Chat Baselines:}
We further benchmark against pathology-trained multimodal models designed for whole-slide reasoning, including SlideChat, MedGemma 1.5, and PathAgent. SlideChat and PathAgent are evaluated using full-WSI inputs following their provided inference scripts. According to the MedGemma instructions, the maximum number of input tiles at 896×896 px is 125. If a slide contains more than 125 patches, tiles are randomly sampled to meet this limit. 

\paragraph{Non-chat Pathology Baselines:}
We evaluate TITAN~\cite{titan_model}, a state-of-the-art pathology foundation model that produces slide-level WSI embeddings optimized for multi-class classification. Because TITAN is not designed for conversational or free-form question answering, we adapt VQA tasks into a multi-class classification format. Specifically, we use GPT-5 to convert each VQA prompt into a structured multiple-choice classification question compatible with TITAN’s inference interface (see Supplementary Material for prompts and examples).
We note that GTEx data were included in TITAN’s pretraining corpus. Results on the GTEx benchmark are therefore labeled accordingly in Table~\ref{tab:all_wsi} to indicate potential pretraining overlap.

\section{Experiments}

\begin{table}[!t]
\centering
\caption{Performance on WSI benchmarks. $\dagger$: Note that TITAN was pretrained on GTEx, while SlideChat and MedGemma were trained on TCGA; this may inflate their performance on benchmarks using these images. Bolded results indicate the best performance among models not specifically trained on the corresponding dataset. Std.\ dev.\ shown in parentheses from 1000 bootstrap replicates.}
\label{tab:all_wsi}
\footnotesize
\setlength{\tabcolsep}{6pt}
\begin{tabularx}{\textwidth}{@{} l l *{5}{>{\centering\arraybackslash}X} @{}}
\toprule
\textbf{Model} & \textbf{Mode} & \textbf{TCGA} & \textbf{GTEx} & \textbf{PANDA} & \textbf{SB VQA} & \textbf{Expert VQA} \\
\midrule

\rowcolor{gray!15}
\multicolumn{7}{l}{\textbf{Pathology-trained WSI Chat Models}} \\
\midrule
SlideChat & WSI 
& 3.3\std{1.2}$^{\dagger}$ 
& 4.7\std{0.3}
& 17.0\std{0.3} 
& \textbf{70.5}\std{3.2} 
& 37.5\std{4.3} \\

MedGemma 1.5 & WSI & 6.2\std{1.5}$^{\dagger}$ & 24.3\std{1.8} & 19.6\std{2.5} & 31.7\std{3.3}$^{\dagger}$ & 35.9\std{4.2}$^{\dagger}$ \\
PathAgent & WSI & 9.3\std{1.7} & 38.9\std{3.2} & 19.2\std{2.5} & 49.5\std{3.6} & 44.5\std{4.5} \\
\midrule

\rowcolor{gray!15}
\multicolumn{7}{l}{\textbf{Pathology-trained ROI Chat Model}} \\
\midrule
\multirow{2}{*}{Patho-R1-7B}
 & Patch     
 & 5.4\std{0.7} 
 & 35.6\std{2.4} 
 & 17.0\std{0.8} 
 & 56.5\std{3.5} 
 & 43.8\std{4.5} \\

 & Thumb 
 & 5.0\std{1.2} 
 & 38.1\std{3.1} 
 & 19.5\std{1.7} 
 & 57.5\std{3.5} 
 & 48.4\std{4.4} \\
\midrule

\multirow{2}{*}{Pathgen-LLaVA}
 & Patch     
 & 8.4\std{1.3} 
 & 50.6\std{2.9} 
 & 17.4\std{0.7} 
 & 52.0\std{3.6} 
 & 42.2\std{4.4} \\

 & Thumb 
 & 4.1\std{1.0} 
 & 34.6\std{2.8} 
 & 19.1\std{1.2} 
 & 50.5\std{3.6} 
 & 38.3\std{4.4} \\
\midrule

\multirow{2}{*}{Quilt-LLaVA}
 & Patch     
 & 4.6\std{0.7} 
 & 25.0\std{2.2} 
 & 16.7\std{0.0} 
 & 37.0\std{3.4} 
 & 31.2\std{4.1} \\

 & Thumb 
 & 3.6\std{0.7} 
 & 11.2\std{2.3} 
 & 8.8\std{1.2} 
 & 33.5\std{3.3} 
 & 35.9\std{4.2} \\
\midrule

\rowcolor{gray!15}
\multicolumn{7}{l}{\textbf{General-purpose Chat Models}} \\
\midrule
\multirow{2}{*}{GPT-5}
 & Patch     
 & 15.5\std{2.2} 
 & 42.3\std{2.4} 
 & 21.1\std{2.4} 
 & 51.5\std{3.6}
 & 41.4\std{4.4} \\

 & Thumb 
 & 9.2\std{1.9} 
 & 36.2\std{3.3} 
 & 12.4\std{2.3} 
 & 54.0\std{3.6} 
 & 50.0\std{4.4} \\
\midrule

\multirow{2}{*}{GPT-4o}
 & Patch     
 & 0.0\std{0.0} 
 & 37.7\std{2.2} 
 & 0.0\std{0.0} 
 & 23.0\std{3.0}
 & 22.7\std{3.6} \\

 & Thumb 
 & 0.0\std{0.0} 
 & 21.5\std{2.6} 
 & 0.0\std{0.0} 
 & 12.0\std{2.2} 
 & 7.0\std{2.3} \\
\midrule

\multirow{2}{*}{Qwen3-VL-235B}
 & Patch     
 & 6.2\std{0.7}
 & 19.8\std{1.3} 
 & 15.2\std{2.0} 
 & 48.5\std{3.4} 
 & 42.2\std{4.3} \\

 & Thumb 
 & 5.6\std{1.0} 
 & 21.0\std{2.0} 
 & 18.7\std{2.2} 
 & 52.5\std{3.6} 
 & 48.4\std{4.4} \\
\midrule

\rowcolor{gray!15}
\multicolumn{7}{l}{\textbf{Non-chat Pathology Models}} \\
\midrule
TITAN & WSI 
& \textbf{88.8\std{1.7}}
& 93.4\std{1.8}$^{\dagger}$
& \textbf{27.0\std{2.4}}
& 39.0\std{3.4} 
& 43.8\std{4.2} \\

\midrule

\rowcolor{gray!15}
\multicolumn{7}{l}{\textbf{GIANT (Training-free WSI Chat Model)}} \\
\midrule
GIANT GPT-5 x5 & WSI 
& 29.3\std{3.3} 
& \textbf{60.5\std{3.3}}
& 25.6\std{2.1} 
& 58.5\std{3.5} 
& \textbf{62.5}\std{4.4} \\

GIANT GPT-5 & WSI 
& 32.3\std{3.5} 
& 54.1\std{3.2} 
& 23.4\std{2.3} 
& 58.0\std{3.5} 
& 57.0\std{4.5} \\

GIANT GPT-4o & WSI 
& 8.5\std{1.5} 
& 42.9\std{3.6} 
& 14.8\std{2.6} 
& 46.5\std{3.6} 
& 40.6\std{4.5} \\

GIANT Qwen3 & WSI 
& 10.5\std{1.6} 
& 26.5\std{2.7} 
& 15.5\std{2.6} 
& 50.0\std{3.6} 
& 53.9\std{4.2} \\

\bottomrule
\end{tabularx}
\end{table}

\subsection{Performance of GIANT and Specialized Pathology Models}

As shown in Table~\ref{tab:all_wsi} and Figure~\ref{fig:results1}, we first found that existing specialist pathology models struggle to generalize to tasks they were not explicitly trained on, particularly large multi-class classification problems such as 30-class cancer diagnosis. On TCGA, SlideChat achieves 3.3\% balanced accuracy, MedGemma 6.2\%, and PathAgent 9.3\%, all near random guessing for a 30-way task. We found that SlideChat performs competitively on its own SlideBenchVQA (e.g., SlideChat 70.5\%), but performance drops substantially on organ classification (4.7\% for SlideChat) and ExpertVQA (37.5\%), highlighting limited cross-task generalization. ROI-based pathology chat models show similarly limited generalizability (e.g., Patho-R1-7B: 5.4\% patch, 5.0\% thumbnail; Pathgen-LLaVA: 8.4\% patch, 4.1\% thumbnail). 

GIANT using GPT-5 substantially outperforms specialized chat models on 4 of 5 tasks (Table~\ref{tab:all_wsi}). On TCGA, GIANT GPT-5 reaches 32.3\% balanced accuracy (29.3\% with five-run aggregation), markedly exceeding all pathology-trained chat baselines. Similar gains are observed on GTEx (54.1\%), PANDA grading (23.4\%), and SlideBenchVQA (58.0\%). On ExpertVQA, GIANT GPT-5 achieves 57.0\%, rising to 62.5\% with five-run aggregation, outperforming all specialist chat models, including SlideChat (37.5\%) and PathAgent (44.5\%). SlideChat remains strongest on SlideBenchVQA, a benchmark derived from the same automated pipeline used for its training data. Examples of model outputs, as well as additional analyses of runtime and cost, are provided in the Supplementary Materials.

\subsection{Pathologist Failure Mode Evaluation}


To understand areas for future improvement, a board-certified pathologist reviewed 28 GIANT reasoning traces on open-ended cancer diagnosis, more challenging than the multi-class TCGA task evaluated here. Failure modes included premature diagnostic anchoring, where the model fixated on an initial hypothesis and reinterpreted contradictory findings to fit it rather than revising the diagnosis. The reviewer also noted recurrent hallucinations of adnexal structures, a tendency to overcall skin-related malignancies, and frequent mislabeling of benign tissue edges as “margins,” which are clinically distinct. Further analysis and examples of reasoning traces are provided in the Supplementary Materials.

\section{Ablation Studies}




\subsection{Text-only Baseline}
To measure the extent to which our benchmarks measure direct reasoning capabilities, we compared GIANT to a text-only baseline using GPT-5. GIANT outperformed the text-only GPT-5 baseline on all tasks, including both SlideBenchVQA and ExpertVQA, with a larger gain on ExpertVQA (+11.7 points) (Table~\ref{tab:textonly}); this is consistent with the design of ExpertVQA, which requires direct slide interpretation.

\begin{table}[h!]
\centering
\caption{Text-only GPT-5 vs.\ GPT-5 GIANT.}
\label{tab:textonly}

\begin{tabularx}{\textwidth}{@{} l *{5}{>{\centering\arraybackslash}X} @{}}
    \toprule
    Method & TCGA & GTEx & PANDA & SB VQA & Expert VQA \\
    \midrule
    Text-only
        & 2.3\std{0.7}
        & 6.2\std{1.8}
        & 5.3\std{1.6}
        & 54.0\std{3.6}
        & 45.3\std{4.4} \\
    GIANT
        & 32.3\std{3.5}
        & 54.1\std{3.2}
        & 23.4\std{2.3}
        & 58.0\std{3.5}
        & 57.0\std{4.5} \\
    \midrule
    $\Delta$ (absolute)
        & +30.0
        & +48.0
        & +18.1
        & +4.0
        & +11.7 \\
    \bottomrule
\end{tabularx}

\end{table}

\subsection{Comparison of Deliberate and Random Region Selection}

First, we conducted a blinded pathologist review of 50 TCGA reasoning traces, comparing GIANT-selected regions against a random baseline. The random baseline uses square bounding boxes containing at least 25\% tissue (via CLAM segmentation). For each reasoning trace, the pathologist was provided the selected region in increments of 5, and asked which set of regions would be more diagnostically useful. As shown in \ref{tab:region_selection_two_panel}A, the pathologist rated GIANT selections to be consistently more diagnostically useful.

We then tested whether this deliberate region selection improves performance over random region selection by replacing GIANT’s navigation with a tool that returns a random region, preserving in-context reasoning. As shown in Table \ref{tab:region_selection_two_panel}B, removing intent caused accuracy to drop across all datasets (e.g., 32.3\% to 26.5\% on TCGA). This demonstrates that intentional cropping of the whole slide image drives GIANT’s gains.

\subsection{Adding a Pathology Foundation Model to GIANT}

A natural extension of an agentic framework, such as GIANT is to incorporate specialist models that provide domain-specific priors. To assess this, we augment GIANT with access to the CONCH pathology model \cite{lu2024avisionlanguage}, enabling the agent to choose at each step between continued navigation or invoking CONCH for localized image–text retrieval. To use this tool, the agent supplies the current crop along with a set of textual hypotheses produced by the LMM. CONCH encodes the image and each hypothesis and returns cosine similarity scores that quantify their alignment, which GIANT can use for subsequent reasoning steps.
Across tasks, adding CONCH leads to minimal change in overall performance, with a notable improvement only on the GTEx organ classification benchmark (Table ~\ref{tab:ablate-conch}). Based on inspection of reasoning traces, we observed that the limited improvement on complex tasks may partly stem from the agent generating incorrect hypotheses, using CONCH to reinforce them.

\begin{table}[h!]
\centering
\caption{\textbf{Ablation: effect of the CONCH tool.}
Std. dev. from 1000 bootstrap replicates.}
\label{tab:ablate-conch}

\begin{tabularx}{\textwidth}{@{} l *{5}{>{\centering\arraybackslash}X} @{}}
    \toprule
    Method & TCGA & GTEx & PANDA & SB VQA & Expert VQA \\
    \midrule
    GIANT
        & 32.3\std{3.5}
        & 54.1\std{3.2}
        & 23.4\std{2.3}
        & 58.0\std{3.5}
        & 57.0\std{4.5} \\
    + CONCH
        & 18.6\std{2.2}
        & 63.7\std{3.1}
        & 25.7\std{2.7}
        & 53.5\std{3.5}
        & 60.9\std{4.5} \\

    \midrule
    $\Delta$ (absolute)
        & --13.7
        & +9.6
        & +2.3
        & --4.5
        & +3.9 \\
    \bottomrule
\end{tabularx}

\end{table}

\begin{table}[t]
\centering
\footnotesize
\caption{\textbf{Intentional vs.\ random region selection.} \textbf{A.} Blinded pathologist evaluation of diagnostically helpful regions. \textbf{B.} Ablation comparing GIANT with intentional region selection vs.\ random region selection (std.\ dev.\ from 1000 bootstrap replicates).}
\label{tab:region_selection_two_panel}

\begin{tabularx}{\textwidth}{@{} l *{3}{>{\centering\arraybackslash}X} @{}}
\toprule
\multicolumn{4}{l}{\textbf{A. Blinded pathologist evaluation (diagnostically helpful regions)}} \\
\midrule
Outcome & GIANT Preferred & Random Preferred & Equally Helpful \\
\midrule
Cases (\%) & 24/31 (77.4) & 7/31 (22.6) & 19/50 (38.0) \\
\bottomrule
\end{tabularx}

\vspace{6pt}

\begin{tabularx}{\textwidth}{@{} l c *{5}{>{\centering\arraybackslash}X} @{}}
\toprule
\multicolumn{7}{l}{\textbf{B. Ablation: intentional region selection vs.\ random}} \\
\midrule
Method & Iters & TCGA & GTEx & PANDA & SB VQA & Expert VQA \\

\midrule
Selected & 20 
    & 32.3\std{3.5}
    & 54.1\std{3.2} 
    & 23.4\std{2.3} 
    & 58.0\std{3.5} 
    & 57.0\std{4.5} \\

Random   & 20 
    & 26.5\std{3.3}
    & 49.4\std{3.3}
    & 19.5\std{2.3}
    & 52.5\std{3.4}
    & 54.7\std{4.4} \\

\midrule
$\Delta$ (absolute)
    & \multicolumn{1}{c}{}
    & +5.8
    & +4.8
    & +3.9
    & +5.5
    & +2.3 \\
\bottomrule
\end{tabularx}
\end{table}




\section{Conclusion}

We presented \textbf{GIANT}, a simple training-free method allowing general-purpose LMMs to navigate gigapixel pathology slides. Beyond the model itself, we introduce \textbf{MultiPathQA} and \textbf{ExpertVQA} to support more realistic and reproducible evaluation of pathology question answering. We show that GPT-5 equipped with GIANT outperforms specialist chat models on a variety of tasks. Together, these findings suggest that iterative multimodal navigation is a promising and practical direction for building more generalizable pathology systems.

\section{Acknowledgments}
We thank Smart In Media for generously offering PathoZoom SlideCloud. We used this as a WSI platform to share cases with pathologists for annotation.

%
%
\bibliographystyle{splncs04}
\bibliography{references}
\end{document}